# A Mask Attention Interaction and Scale Enhancement Network for SAR Ship Instance Segmentation

Tianwen Zhang, and Xiaoling Zhang

*Abstract*—**Most of existing synthetic aperture radar (SAR) ship instance segmentation models do not achieve mask interaction or offer limited interaction performance. Besides, their multi-scale ship instance segmentation performance is moderate especially for small ships. To solve these problems, we propose a mask attention interaction and scale enhancement network (MAI-SE-Net) for SAR ship instance segmentation. MAI uses an atrous spatial pyramid pooling (ASPP) to gain multi-resolution feature responses, a non-local block (NLB) to model long-range spatial dependencies, and a concatenation shuffle attention block (CSAB) to improve interaction benefits. SE uses a content-aware reassembly of features block (CARAFEB) to generate an extra pyramid bottom-level to boost small ship performance, a feature balance operation (FBO) to improve scale feature description, and a global context block (GCB) to refine features. Experimental results on two public SSDD and HRSID datasets reveal that MAI-SE-Net outperforms the other nine competitive models, better than the suboptimal model by 4.7% detection AP and 3.4% segmentation AP on SSDD and by 3.0% detection AP and 2.4% segmentation AP on HRSID.**

## I. INTRODUCTION

SYNTHETIC aperture radar (SAR) is widely used for ocean ship monitoring. Classic methods extract features using expert experience to search for ships [1], [2], e.g., constant false alarm rate (CFAR) [3] and template-based matching [4]. Modern methods use convolutional neural networks (CNNs) [5], [6] to extract features with less human involvement, receiving more attention in SAR community. Most scholars still used rectangular bounding boxes [7]–[11] to detect ships; SAR ship segmentation at the box- and pixel-level received less attention.

Yet, ship instance segmentation is important, which cannot be ignored. This is because it can search every pixel of the ship entity, conducive to all-around ocean monitoring, e.g., ship contour extraction, ship category recognition [12], etc. Moreover, a rotating bounding box [13] can considerably improve the detection performance of object in remote sensing images, but it still cannot achieve refined pixel-level object segmentation. In fact, if a ship instance is segmented successfully, its rotating bounding box can be obtained, i.e., the minimum circumscribed rectangle of a segmentation polygon. Thus, ship instance segmentation is a higher-level task than rotated ship detection.

A small amount of reports conducted SAR ship instance segmentation [14]–[21]. Wei *et al.* [14] released the HRSID dataset used for SAR ship instance segmentation, but no methodological contributions were offered. Su *et al.* [15] proposed an HQ-

This work was supported by the National Natural Science Foundation of China under Grant 61571099. *(Corresponding author: Xiaoling Zhang.)*

The authors are with the School of Information and Communication Engineering, University of Electronic Science and Technology of China, Chengdu 611731, China (e-mail: twzhang@std.uestc.edu.cn; xlzhang@uestc.edu.cn)

ISNet model to segment targets in remote sensing images, but they did consider ship characteristics. Zhao *et al.* [16] designed a synergistic attention for better accuracy, but there were still many missed detections in complex scenes. Gao *et al.* [17] proposed an anchor-free model with a centroid distance loss to enhance performance, but their models still lacks the capacity to deal with complex scenes and cases. Zhang *et al.* [18] used the hybrid task cascade (HTC) [19] for SAR ship instance segmentation, yielding better performance than Mask R-CNN [20]. Fan *et al.* [21] used the swin-transformer to extract ship features, but their mask prediction accuracy is still limited. Especially, most of these existing studies do not achieve mask interaction or offer limited interaction benefits, hampering mask prediction performance improvement. Besides, their multi-scale instance segmentation performance is moderate. For one thing, many small ships are often missed. For another, the pixel segmentation of large ships is not accurate enough.

To solve these problems, this letter proposes a mask attention interaction and scale enhancement network (MAI-SE-Net) for better SAR ship instance segmentation. For better mask interaction, based on HTC [19], MAI adopts an atrous spatial pyramid pooling (ASPP) to excite multi-resolution feature responses, a non-local block (NLB) to establish long-range spatial dependencies, and a concatenation shuffle attention block (CSAB) to boost interaction benefits. For better multi-scale performance, based on FPN [22], SE employs a content-aware reassembly of features block (CARAFEB) to generate an extra pyramid bottom-level to enhance small ship performance, a feature balance operation (FBO) to boost multi-scale feature description, and a global context block (GCB) to refine features. We conduct experiments on two open SSDD [23] and HRSID [14] datasets, and the results indicate the state-of-the-art performance of the proposed MAI-SE-Net, compared with the other nine competitive models. Specifically, MAI-SE-Net is superior to the suboptimal model by 4.7% detection average precision (AP) and 3.4% segmentation AP on SSDD, and by 3.0% detection AP and 2.4% segmentation AP on HRSID. The ablation studies verify each technique's effectiveness in MAI and SE.

The main contributions of this letter are as follows.

1) MAI-SE-Net is established for better SAR ship instance segmentation. It can enable enhanced mask interaction benefits and better multi-scale performance.
2) ASPP, NLB and CSAB are inserted into the mask interaction network to enhance interaction gain.
3) CARAFEB, FBO and GCB are inserted into FPN for better multi-scale performance especially for small ships.

The rest of this letter is arranged as follows. Section II introduces the proposed MAI-SE-Net. Section III describes our experiments. The results are shown in Section IV. Discussions are given in Section V. Finally, a summary is made in Section VI.

## II. MAI-SE-NET

Fig. 1 shows MAI-SE-Net's overall framework that follows the two-stage paradigm [20] for instance segmentation. A backbone network is used to extract features. SE can enhance multiscale performance of FPN by CARAFEB, FBO, and GCB. A region proposal network (RPN) [24] is used to generate regions of interests (ROIs), which are mapped to feature maps to extract feature subsets by ROIAlign [20]. MAI is to boost mask interaction performance of HTC by ASPP, NLB and CSAB. Next, we will introduce the implementations of MAI and SE in detail.

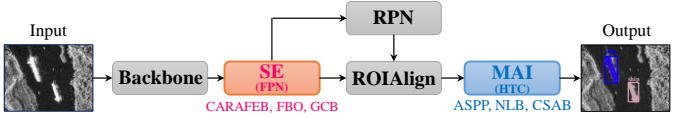

Fig. 1. Overall framework of MAI-SE-Net.

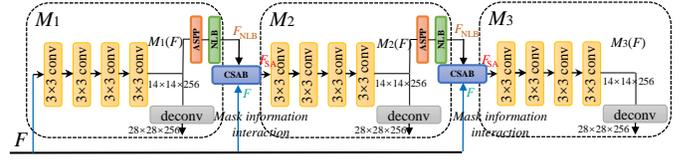

Fig. 2. Implementation of the mask attention interaction (MAI).

### A. Mask Attention Interaction (MAI)

A vital novelty of HTC is the multi-stage mask branches [19]. It offers mask information interaction with better performance. MAI boosts mask information interaction benefits further. Fig. 2 is MAI's implementation. It has three prediction branches $M_1$, $M_2$, $M_3$ trained with increasing intersection over union (IOU) thresholds. Features of previous stage $M_{i-1}$ are refined by ASPP and NLB for next stage $M_i$. CSAB is used for feature fusion.

*1) ASPP.* SAR ships contain various context surroundings as in Fig. 3(a), e.g., blur edge, sidelobe, wake, speckle noise, tower crane [25] (i.e., a self-contained metallic equipment for loading and unloading cargo on a ship), and inshore facilities. Adding contexts is conducive to easing background interferences. This inspires us to adopt atrous convs [26] with different rates to obtain multi-resolution feature responses, so as to ensure ample receptive fields of ship surroundings. We apply ASPP [27] designed for semantic segmentation to reach it, because its advanced performance has been verified in pixel-sensitive tasks. Fig. 4(a) shows its implementation, described by

$$F_{ASPP} = f_{1 \times 1}\{\{f_{3 \times 3}^2(M_i(F)), f_{3 \times 3}^3(M_i(F)), f_{3 \times 3}^4(M_i(F)), f_{3 \times 3}^5(M_i(F))\}\} \quad (1)$$

where $f^r_{3 \times 3}$ means a $3 \times 3$ conv with a dilated rate of $r$, and $f_{3 \times 3}$ is a $1 \times 1$ conv for channel reduction ($4C \rightarrow C$). Different dilated rates enable different resolution responses, with different range contexts. We set four dilated rates for the accuracy-speed trade-off. More may offer better performance but must sacrifice speed. Different from [27], the four dilated rates are set to 2, 3, 4, and 5, due to the small size of mask feature maps ($14 \times 14$).

*2) NLB.* ASPP adopts four parallel $3 \times 3$ convs to extract features with different range contexts, but it may import excessive contexts potentially when the dilated rate comes too big. In this case, extravagant background clutters may dominate networks, leading to inaccurate positionings. We use NLB [28] to avoid it.

Zhang *et al.* [28] gives generic theories of non-local networks,

$$\mathbf{y}_i = \frac{1}{\zeta(\mathbf{x})} \sum_{\forall j} f(\mathbf{x}_i, \mathbf{x}_j) g(\mathbf{x}_j) \quad (2)$$

where $\mathbf{x}$ is NLB's input, i.e. $F_{APSS}$, $i$ and $j$ are the index position of input feature maps across the $H \times W$ space. Here, $H$ and $W$ denote the height and width of feature maps. $f$ is to represent spatial correlation between $i$ and all $j$, thus the $i$-position's output $\mathbf{y}_i$ is related to the entire space. Global long-range spatial dependencies are captured. $g$ denotes the representation of the input at the $j$-position. $f \cdot g$ means the resulting response and then it is normalized by a factor $\zeta(\mathbf{x})$.

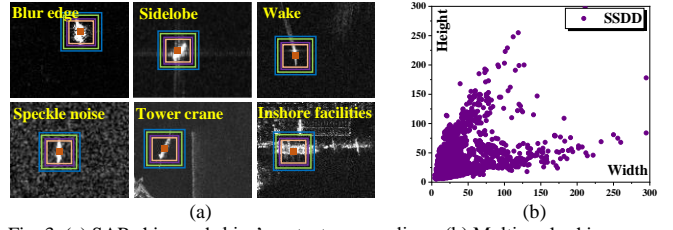

Fig. 3. (a) SAR ships and ships' context surroundings. (b) Multi-scale ships.

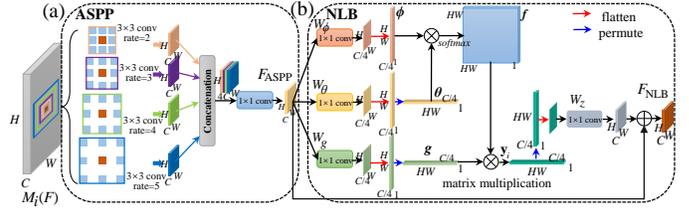

Fig. 4. (a) Implementation of ASPP. (b) Implementation of NLB.

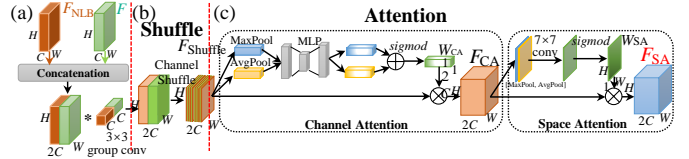

Fig. 5. Implementation of CSAB. (a) Concatenation. (b) Shuffle. (c) Attention.

We instantiate (2) in Fig. 4(b). Features at the $i$-position are denoted by $\phi$ using a $1 \times 1$ conv $W_\phi$. Features at the $j$-position are denoted by $\theta$ using a $1 \times 1$ conv $W_\theta$. $f$ is obtained from adaptive learning between $\phi$ and $\theta$ where the normalization process is equivalent to a *softmax* calculation function [28]. The representation of the input at the $j$-position $g$ is learned using another one $1 \times 1$ conv $W_g$. The response at the $i$-position $\mathbf{y}_i$ is obtained by a matrix multiplication. Note that we embed all features into $C/4$ channel space to focus on the most important quarter contexts from the raw $4C$ channel space. This can also reduce computational burdens. To apply response to the input readily, we use another one $1 \times 1$ conv $W_z$ to transform dimension for the adding operation. Finally, we obtain the final feature self-attention output $F_{NLB}$. Readers can consult more details in [28].

*3) CSAB.* Three steps constitute CSAB as in Fig. 5, i.e., concatenation, shuffle, and attention. We follow the feature reuse idea to concatenate feature maps $F_{NLB}$ and $F$ from the previous stage $M_{i-1}$ and the backbone network. A $3 \times 3$ group conv is to refine them separately. We then shuffle the obtained results in the channel dimension to reduce channel synergy consistency, and the result is $F_{Shuffle}$. Finally, we use CBAM [29] for a channel and space attention. $W_{CA}$ is the channel attention weight and $W_{SA}$ is the space attention weight. They range from 0 to 1 by a *sigmod* activation. The terminal output of CSAB denotes $F_{SA} = F_{Shuffle} \cdot W_{CA} \cdot W_{SA}$ that will be transmitted to the next stage $M_i$.

### B. Scale Enhancement (SE)

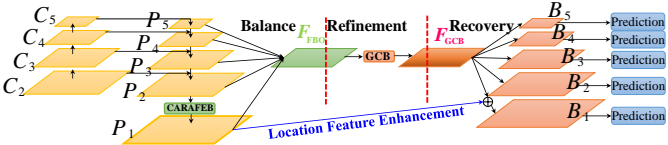

Fig. 6. Implementation of the scale enhancement (SE).

Fig. 6 is SE's implementation. We use CARAFE to generate a bottom-level $P_1$ to boost small ship performance. FBO is used to improve scale feature description. GCB is used to refine features. Finally, we reconstruct FPN for multi-scale prediction.

*1) CARAFEB.* Different from natural objects in optical images, SAR is a "bird-eye" remote sensing satellite, thus most SAR ships are rather small [23] as in Fig. 3(b). However, the original FPN was designed for natural objects which is not good at detecting small ships. We advocate to generate an extra FPN bottom-level $P_1$ to enhance small ship instance performance. CARAFEB [30] is suggested because it has better performance than bilinear interpolation and deconv. It enables instance-specific content-aware handling to generate adaptive kernels on-the-fly. Readers can find it detailed implementation in [30], [31].

*2) FBO.* Combined with CARAFEB, FPN contains five levels ($P_1$, $P_2$, $P_3$, $P_4$ and $P_5$). We observe that more levels might bring unstable training due to the huge feature gap between $P_1$ and $P_5$, so a feature balance operation is needed [32]. We perform it at the middle $P_3$ level due to its all-round semantic and location features. The above is described by

$$F_{FBO} = [\text{MP}^{\times 4}(P_1) + \text{MP}^{\times 2}(P_2) + P_3 + \text{US}^{\times 2}(P_4) + \text{US}^{\times 4}(P_5)] / 5 \quad (3)$$

where $\text{MP}^{\times n}$ means $n$ times maxpooling and $\text{US}^{\times n}$ means $n$ times up-sampling. In this way, ship features with more balanced spatial information and semantic information can be obtained.

*3) GCB.* We also adopt the advanced GCB [33] to refine features further for a more robust FPN. Pang *et al.* [34] used a non-local block for feature refinement, but they did not consider the balance of different levels. GCB can meet the non-local and squeeze-excitation concurrently, with better performance [33]. Readers can find its detailed implementation in [33].

Finally, we use the opposite operation of (3) to recover the final FPN ($B_1$, $B_2$, $B_3$, $B_4$ and $B_5$) for prediction. Especially, to avoid the possible loss of the low-level location features, we also add the raw $P_1$ to the results of GCB when recovering $B_1$.

## III. Experiments

### A. Dataset

SSDD [23] has 1160 samples with 512×512 size from Radar-Sat-2, TerraSAR-X and Sentinel-1. The training-test ratio is 4:1 [23]. HRSID [14] has 5604 samples with 800×800 average size from TerraSAR-X and Sentinel-1. The training set has 3643 samples and the rest serves as the test set, same as [14].

### B. Training Details

ResNet-101 [35] pretrained on ImageNet [36] serves as backbone networks. We train all networks by 12 epochs using SGD [37] whose learning rate is 0.002 reduced 10 times at 8- and 11-epoch. The batch size is 4. In inference, NMS [38] removes redundant detections. Experiments run on a computer with 3090 GPU and i9 CPU with MMDetection and Pytorch.

### C. Evaluation Criteria

We adopt the COCO evaluation criteria [18]. AP is the average precision of different IOU thresholds with 0.50:0.05:0.95. $AP_{50}$ is the accuracy with a 0.50 IOU threshold. $AP_{75}$ is that with a 0.75 IOU threshold. $AP_S$ is that of small ships ($< 32^2$ pixels). $AP_M$ is that of medium ships ($> 32^2$ pixels and $< 96^2$ pixels). $AP_L$ is that of large ships ($> 96^2$ pixels).

## IV. Results

### A. Quantitative Results

TABLE I-II exhibit the quantitative results on SSDD and HRSID. MAI-SE-Net offers the best accuracy, better than the second-best model by 4.7% detection AP and 3.4% segmentation AP on SSDD; by 3.0% detection AP and 2.4% segmentation AP on HRSID. This shows the advanced performance of MAI-SE-Net. $AP_L$ has some fluctuations because large ships are rather few as in Fig. 3(b). Each technique, including ASPP, NLB and CSAB in MAI, and CARAFEB, FBO and GCB in SE, offer an observable accuracy increase. This shows their effectiveness. The speed of MAI-SE-Net is modest, because the calculation cost is increased. However, it should be acceptable, because it is still faster than HQ-ISNet. Though YOLACT's speed is highest, its accuracy is too poor to meet applications.

TABLE I
Quantitative Results on SSDD. The Suboptimal Method Is Marked by Underline "—".

| Mask Attention Interaction | | | Scale Enhancement | | | Detection Task (%) | | | | | | Segmentation Task (%) | | | | | | FPS |
|---|---|---|---|---|---|---|---|---|---|---|---|---|---|---|---|---|---|---|
| ASPP | NLB | CSAB | CARAFEB | FBO | GCB | AP | $AP_{50}$ | $AP_{75}$ | $AP_S$ | $AP_M$ | $AP_L$ | AP | $AP_{50}$ | $AP_{75}$ | $AP_S$ | $AP_M$ | $AP_L$ | |
| -- | -- | -- | -- | -- | -- | 65.6 | 93.6 | 76.3 | 65.2 | 68.4 | 27.5 | 59.3 | 91.7 | 73.1 | 58.7 | 61.6 | 34.8 | 11.60 |
| ✓ | | | | | | 66.0 | 93.0 | 76.6 | 65.8 | 68.7 | 39.6 | 59.9 | 92.0 | 73.9 | 59.3 | 62.1 | 60.2 | 11.05 |
| ✓ | ✓ | | | | | 66.8 | 93.4 | 79.8 | 66.4 | 69.9 | 26.0 | 60.8 | 92.2 | 75.4 | 60.2 | 63.2 | 40.1 | 10.55 |
| ✓ | ✓ | ✓ | | | | 67.8 | 93.3 | 80.2 | 67.5 | 71.0 | 46.2 | 61.1 | 92.1 | 76.1 | 60.6 | 62.4 | 55.2 | 10.09 |
| ✓ | ✓ | ✓ | ✓ | | | 68.1 | 96.2 | 80.3 | 68.1 | 68.6 | 33.3 | 62.1 | 94.3 | 76.2 | 62.1 | 62.1 | 45.2 | 9.28 |
| ✓ | ✓ | ✓ | ✓ | ✓ | | 69.4 | 95.3 | 83.4 | 69.5 | 69.9 | 13.3 | 62.5 | 93.4 | 77.5 | 62.5 | 62.6 | 17.7 | 8.92 |
| ✓ | ✓ | ✓ | ✓ | ✓ | ✓ | 70.3 | 96.2 | 85.3 | 70.7 | 70.9 | 37.7 | 63.0 | 94.4 | 77.6 | 63.3 | 62.5 | 47.7 | 8.29 |
| | | | | | | +4.7 | +2.6 | +9.0 | +5.0 | +2.5 | +10.2 | +3.7 | +2.7 | +4.5 | +4.6 | +0.9 | +12.9 | |
| Mask R-CNN [20] | | | ResNet-101-FPN | | | 62.0 | 91.5 | 75.4 | 62.0 | 64.4 | 19.7 | 57.8 | 88.5 | 72.1 | 57.2 | 60.8 | 27.4 | 11.05 |
| Mask Scoring R-CNN [39] | | | ResNet-101-FPN | | | 62.4 | 91.0 | 75.1 | 61.9 | 66.0 | 15.7 | 58.6 | 89.4 | 73.2 | 58.0 | 61.4 | 22.6 | 12.88 |
| Cascade Mask R-CNN [40] | | | ResNet-101-FPN | | | 63.0 | 89.6 | 75.2 | 62.4 | 66.0 | 12.0 | 56.6 | 87.5 | 70.5 | 56.3 | 58.8 | 22.6 | 10.55 |
| HTC [19] | | | ResNet-101-FPN | | | 65.6 | 93.6 | 76.3 | 65.2 | 68.4 | 27.5 | 59.3 | 91.7 | 73.1 | 58.7 | 61.6 | 34.8 | 11.60 |
| PANet [41] | | | ResNet-101-FPN | | | 63.3 | 93.4 | 75.4 | 63.4 | 65.5 | 40.8 | 59.6 | 91.1 | 74.0 | 59.3 | 61.0 | 52.1 | 13.65 |
| YOLACT [42] | | | ResNet-101-FPN | | | 54.0 | 90.6 | 61.2 | 56.9 | 48.2 | 12.6 | 48.4 | 88.0 | 52.1 | 47.3 | 53.5 | 40.2 | 15.47 |
| GRoIE [43] | | | ResNet-101-FPN | | | 61.2 | 91.5 | 71.6 | 62.2 | 59.8 | 8.7 | 58.3 | 89.8 | 72.7 | 58.6 | 58.7 | 21.8 | 9.67 |
| HQ-ISNet [15] | | | HRNetV2-W18 | | | 64.9 | 91.0 | 76.3 | 64.7 | 66.6 | 26.0 | 58.6 | 89.3 | 73.6 | 58.2 | 60.4 | 37.2 | 8.59 |
| HQ-ISNet [15] | | | HRNetV2-W32 | | | 65.5 | 90.7 | 77.3 | 65.6 | 66.9 | 23.2 | 59.3 | 90.4 | 75.5 | 58.9 | 61.1 | 37.3 | 8.00 |
| HQ-ISNet [15] | | | HRNetV2-W40 | | | 63.6 | 87.8 | 75.3 | 62.6 | 67.8 | 27.9 | 57.6 | 86.0 | 72.6 | 56.7 | 61.3 | 50.2 | 7.73 |
| SA R-CNN [16] | | | ResNet-50-GCB-FPN | | | 63.2 | 92.1 | 75.2 | 63.8 | 64.0 | 7.0 | 59.4 | 90.4 | 73.3 | 59.6 | 60.3 | 20.2 | 13.65 |
| **MAI-SE-Net (Ours)** | | | ResNet-101-FPN | | | 70.3 | 96.2 | 85.3 | 70.7 | 70.9 | 37.7 | 63.0 | 94.4 | 77.6 | 63.3 | 62.5 | 47.7 | 8.29 |
| | | | | | | +4.7 | +2.6 | +8.0 | +5.1 | +4.3 | -3.1 | +3.4 | +2.7 | +2.1 | +3.7 | +0.9 | -4.4 | |

TABLE II
QUANTITATIVE RESULTS ON HRSID. THE SUBOPTIMAL METHOD IS MARKED BY UNDERLINE "——".

| Mask Attention Interaction | | | Scale Enhancement | | | Detection Task (%) | | | | | | Segmentation Task (%) | | | | | | FPS |
|---|---|---|---|---|---|---|---|---|---|---|---|---|---|---|---|---|---|---|
| ASPP | NLB | CSAB | CARAFEB | FBO | GCB | AP | AP50 | AP75 | APS | APM | APL | AP | AP50 | AP75 | APS | APM | APL | |
| -- | -- | -- | -- | -- | -- | 66.6 | 86.0 | 77.1 | 67.6 | 69.0 | 28.1 | 55.2 | 84.9 | 66.5 | 54.7 | 63.8 | 19.2 | 7.42 |
| ✓ | | | | | | 67.4 | 88.0 | 78.3 | 68.3 | 69.5 | 24.7 | 55.9 | 86.8 | 67.2 | 55.3 | 63.8 | 19.0 | 7.10 |
| ✓ | ✓ | | | | | 67.9 | 88.2 | 78.6 | 68.7 | 70.8 | 25.9 | 56.3 | 86.3 | 67.3 | 55.8 | 64.4 | 18.8 | 6.75 |
| ✓ | ✓ | ✓ | | | | 68.4 | 89.9 | 78.6 | 69.6 | 69.8 | 27.0 | 56.5 | 87.9 | 68.1 | 56.1 | 63.9 | 21.7 | 6.40 |
| ✓ | ✓ | ✓ | ✓ | | | 68.8 | 90.9 | 78.7 | 70.0 | 67.9 | 23.5 | 57.0 | 88.1 | 68.1 | 56.5 | 63.7 | 21.6 | 5.98 |
| ✓ | ✓ | ✓ | ✓ | ✓ | | 69.2 | 90.9 | 78.8 | 70.4 | 69.4 | 29.5 | 57.3 | 88.8 | 68.1 | 57.1 | 62.5 | 23.5 | 5.61 |
| ✓ | ✓ | ✓ | ✓ | ✓ | ✓ | 69.7 | 91.9 | 79.6 | 70.7 | 69.4 | 34.0 | 57.8 | 89.2 | 68.6 | 57.6 | 63.0 | 24.4 | 5.30 |
| | | | | | | +3.1 | +5.9 | +2.5 | +3.1 | +0.4 | +5.9 | +2.6 | +4.3 | +2.1 | +2.9 | -0.8 | +5.2 | |
| Mask R-CNN [20] | ResNet-101-FPN | | | | | 65.1 | 87.7 | 75.5 | 66.1 | 68.4 | 14.1 | 54.8 | 85.7 | 65.2 | 54.3 | 62.5 | 13.3 | 7.07 |
| Mask Scoring R-CNN [39] | ResNet-101-FPN | | | | | 65.2 | 87.6 | 75.4 | 66.5 | 67.4 | 13.4 | 54.9 | 85.1 | 65.9 | 54.5 | 61.5 | 12.9 | 8.24 |
| Cascade Mask R-CNN [40] | ResNet-101-FPN | | | | | 65.1 | 85.4 | 74.4 | 66.0 | 69.0 | 17.1 | 52.8 | 83.4 | 62.9 | 52.2 | 62.2 | 17.0 | 6.75 |
| HTC [19] | ResNet-101-FPN | | | | | 66.6 | 86.0 | 77.1 | 67.6 | 69.0 | 28.1 | 55.2 | 84.9 | 66.5 | 54.7 | 63.8 | 19.2 | 8.74 |
| PANet [41] | ResNet-101-FPN | | | | | 65.4 | 88.0 | 75.7 | 66.5 | 68.2 | 22.1 | 55.1 | 86.0 | 66.2 | 54.7 | 62.8 | 17.8 | 8.74 |
| YOLACT [42] | ResNet-101-FPN | | | | | 47.9 | 74.4 | 53.3 | 51.7 | 34.9 | 3.3 | 39.6 | 71.1 | 41.9 | 39.5 | 46.1 | 7.3 | 10.02 |
| GRoIE [43] | ResNet-101-FPN | | | | | 65.4 | 87.8 | 75.5 | 66.5 | 67.2 | 21.8 | 55.4 | 85.8 | 66.9 | 54.9 | 63.5 | 19.7 | 6.19 |
| HQ-ISNet [15] | HRNetV2-W18 | | | | | 66.0 | 86.1 | 75.6 | 67.1 | 66.3 | 8.9 | 53.4 | 84.2 | 64.3 | 53.2 | 59.7 | 10.7 | 5.50 |
| HQ-ISNet [15] | HRNetV2-W32 | | | | | 66.7 | 86.9 | 76.3 | 67.8 | 68.3 | 16.8 | 54.6 | 85.0 | 65.8 | 54.2 | 61.7 | 13.4 | 5.12 |
| HQ-ISNet [15] | HRNetV2-W40 | | | | | 66.7 | 86.2 | 76.3 | 67.9 | 68.6 | 11.7 | 54.2 | 84.3 | 64.9 | 53.9 | 61.9 | 12.8 | 4.95 |
| SA R-CNN [16] | ResNet-50-GCB-FPN | | | | | 65.2 | 88.3 | 75.2 | 66.4 | 65.4 | 10.2 | 55.2 | 86.2 | 66.7 | 54.9 | 60.9 | 12.3 | 8.74 |
| **MAI-SE-Net (Ours)** | ResNet-101-FPN | | | | | 69.7 | 91.9 | 79.6 | 70.7 | 69.4 | 34.0 | 57.8 | 89.2 | 68.6 | 57.6 | 63.0 | 24.4 | 5.30 |
| | | | | | | +3.0 | +3.9 | +2.5 | +2.8 | +0.4 | +5.9 | +2.4 | +3.0 | +1.7 | +2.7 | -0.8 | +5.2 | |

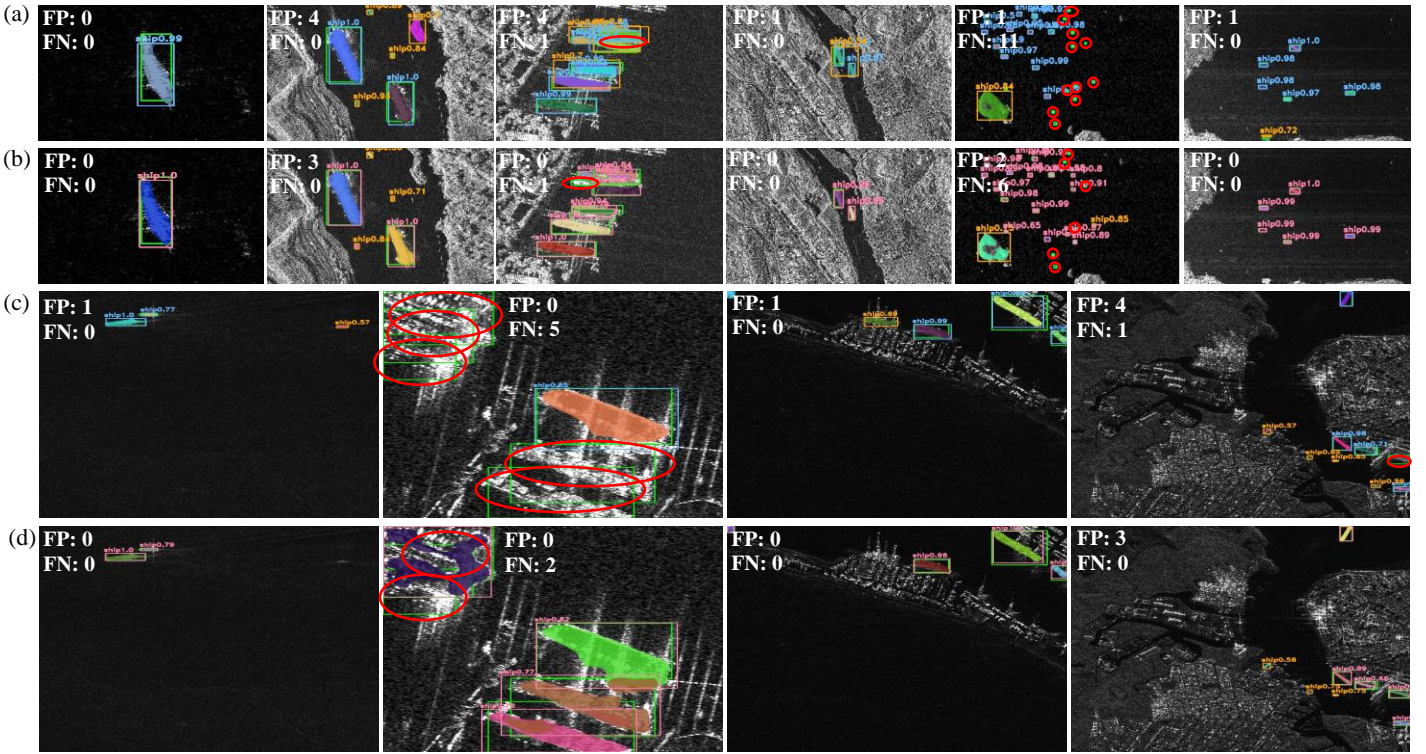

Fig. 7. Qualitative results. (a) Suboptimal HTC on SSDD. (b) Our MAI-SE-Net on SSDD. (c) Suboptimal HTC on HRSID. (b) Our MAI-SE-Net on HRSID. Green boxes denote ground truths offered by data releasers. Orange boxes denote false alarms (false positives, FP). Red ellipses denote missed detections (false negatives, FN). The colored-pixel indeed means pixel-level result and the colored-box means target-level result. The value above the box indicates ship confidence.

## B. Qualitative Results

The qualitative results on two datasets are shown in Fig. 7. Here, we only show the qualitative comparison results with the suboptimal model for limited pages. From Fig. 7, MAI-SE-Net detected more ships with high confidence, generated less false alarms (false positives, FP), and avoided less missed detections (false negatives, FN). For example, for the first sample in Fig. 7(a)(b), the ship confidence of MAI-SE-Net is 1.0, greater than HTC's 0.99; for the second sample in Fig. 7(a)(b), MAI-SE-Net generates three false alarms, fewer than HTC's four ones; for fifth sample in Fig. 7(a)(b), MAI-SE-Net has six missed detections, fewer than HTC's eleven ones. The above reveals MAI-

SE-Net's superiority. Moreover, many small ships are detected by MAI-SE-Net, benefiting from our proposed improvements.

## V. DISCUSSIONS

We survey the contribution of each technique in MAI and SE on the overall performance in TABLE III–VIII, i.e., remove and then install each technique separately. From TABLE III–VIII, each technique in MAI and SE can improve performance individually, which further confirms their effectiveness. This is in line with the incremental increase in accuracy when using more and more techniques in TABLE I–II. Moreover, the sensitivities of MAI-SE-Net to different techniques are different, e.g., FBO

TABLE III
QUANTITATIVE RESULTS WITH AND WITHOUT ASPP ON SSDD.

| Case | Detection Task (%) | | | | | | Segmentation Task (%) | | | | | | FPS |
|---|---|---|---|---|---|---|---|---|---|---|---|---|---|
| | AP | AP$_{50}$ | AP$_{75}$ | AP$_S$ | AP$_M$ | AP$_L$ | AP | AP$_{50}$ | AP$_{75}$ | AP$_S$ | AP$_M$ | AP$_L$ | |
| Case 1 | 69.5 | 96.1 | 82.3 | 69.4 | 70.9 | 68.3 | 62.7 | 94.5 | 77.6 | 62.9 | 62.8 | 50.0 | 8.55 |
| Case 2 | 69.7 | 95.3 | 84.1 | 69.9 | 70.2 | 50.9 | 62.7 | 93.4 | 76.4 | 63.3 | 61.9 | 50.1 | 8.90 |
| Case 3 | 70.1 | 96.2 | 83.9 | 70.4 | 69.8 | 56.1 | 62.8 | 94.4 | 78.9 | 62.9 | 62.6 | 52.5 | 8.47 |
| Case 4 | 68.7 | 96.1 | 80.7 | 68.4 | 70.6 | 56.8 | 61.5 | 94.3 | 76.3 | 60.9 | 63.6 | 62.5 | 9.46 |
| Case 5 | 68.4 | 95.8 | 82.0 | 69.3 | 67.5 | 50.0 | 61.6 | 94.4 | 75.8 | 61.1 | 63.4 | 60.0 | 8.72 |
| Case 6 | 69.4 | 95.3 | 83.4 | 69.5 | 69.9 | 13.3 | 62.5 | 93.4 | 77.5 | 62.5 | 62.6 | 17.7 | 8.92 |
| MAI-SE-Net | **70.3** | 96.2 | 85.3 | 70.7 | 70.9 | 37.7 | **63.0** | 94.4 | 77.6 | 63.3 | 62.5 | 47.7 | 8.29 |

Case 1 means that MAI-SE-Net only removes ASPP. Case 2 means that MAI-SE-Net only removes NLB. Case 3 means that MAI-SE-Net only removes CSAB. Case 4 means that MAI-SE-Net only removes CARAFEB. Case 5 means that MAI-SE-Net only removes FBO. Case 6 means that MAI-SE-Net only removes GCB.

offers a 1.9% detection AP gain and a 1.4% segmentation AP gain, while the accuracy gains are 0.2% and 0.2% for CSAB. This needs to be studied in the future.

## VI. CONCLUSION

We propose MAI-SE-Net for better SAR ship instance segmentation. MAI uses ASPP, NLB, and CSAB to enhance mask interaction performance. SE uses CARAFEB, FBO, and GCB to improve multi-scale prediction performance. Results on two open datasets show the state-of-the-art performance of the proposed MAI-SE-Net, better than the other nine competitive models. Finally, the ablation studies confirm the effectiveness of each technique in MAI and SE. In the future, we will optimize the inference speed of MAI-SE-Net.